\documentclass[10pt,twocolumn,letterpaper]{article}

\usepackage[pagenumbers]{cvpr} 

\usepackage[accsupp]{axessibility}  

\usepackage{graphicx}
\usepackage{amsmath}
\usepackage{amssymb}
\usepackage{booktabs}
\usepackage[table]{xcolor}
\usepackage{comment}
\usepackage{enumitem}
\usepackage{float}

\usepackage[pagebackref,breaklinks,colorlinks]{hyperref}

\usepackage[capitalize]{cleveref}
\crefname{section}{Sec.}{Secs.}
\Crefname{section}{Section}{Sections}
\Crefname{table}{Table}{Tables}
\crefname{table}{Tab.}{Tabs.}

\usepackage[linesnumbered,ruled,vlined,nofillcomment]{algorithm2e}
\SetKwInput{KwInput}{Input}
\SetKwInput{KwOutput}{Output}
\SetKwInput{KwInit}{Init}
\SetKwFor{Find}{find}{do}{}
\SetKwFunction{KwShap}{Shapely}
\SetKwProg{Fn}{Function}{:}{\KwRet}

\def\cvprPaperID{34} 
\def\confName{XAI4CV}
\def\confYear{2022}

\begin{document}

\title{Exploring Concept Contribution Spatially: \\
Hidden Layer Interpretation with Spatial Activation Concept Vector}

\author{Andong Wang, Wei-Ning Lee\\
The University of Hong Kong\\
{\tt\small wangad@connect.hku.hk, \tt\small wnlee@eee.hku.hk}
}





\twocolumn[{%
\maketitle
\begin{figure}[H]
    \hsize=\textwidth
    \centering
    \includegraphics[width=16.9cm]{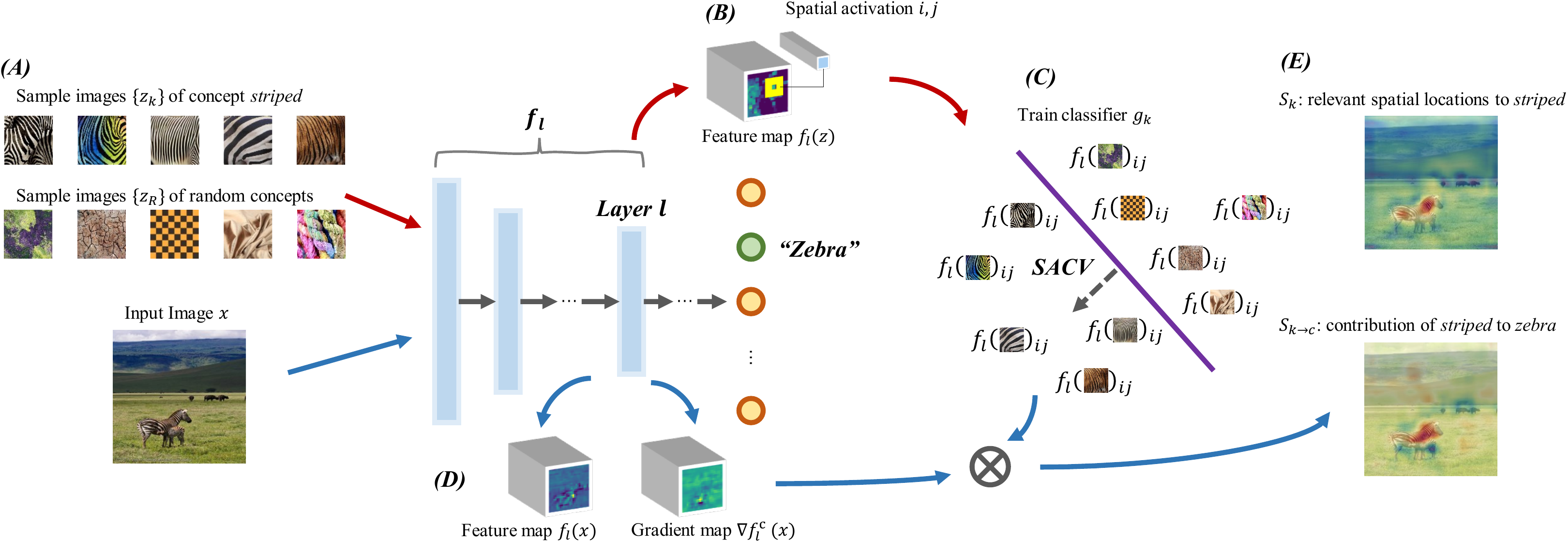}
    \setlength{\belowcaptionskip}{0pt}
    \caption{Pipeline of \textbf{Spatial Activation Concept Vector (SACV)}. See Sec. \ref{sec:method} for details of each step.}
    \label{fig:pipeline}
\end{figure}
}]

\section{Introduction}

To interpret deep learning models, one mainstream is to explore the learned concepts by networks \cite{kim2018interpretability, olah2018feature, bau2020understanding}. \emph{Testing with Concept Activation Vector (TCAV)} \cite{kim2018interpretability} presents a powerful tool to quantify the contribution of query concepts (represented by user-defined guidance images) to a target class. For example, we can quantitatively evaluate whether and to what extent concept \emph{striped} contributes to model prediction \emph{zebra} with TCAV. Therefore, TCAV whitens the reasoning process of deep networks. And it has been applied to solve practical problems such as diagnosis \cite{cai2019human}. 

However, for some images where the target object only occupies a small fraction of the region, TCAV evaluation may be interfered with by redundant background features because TCAV calculates concept contribution to a target class based on a whole hidden layer. 

To tackle this problem, based on TCAV, we propose \textbf{Spatial Activation Concept Vector (SACV)} which identifies the relevant spatial locations to the query concept while evaluating their contributions to the model prediction of the target class. Experiment shows that SACV generates a more fine-grained explanation map for a hidden layer and quantifies concepts' contributions spatially. Moreover, it avoids interference from background features. Code is available on \url{https://github.com/AntonotnaWang/Spatial-Activation-Concept-Vector}.

\begin{figure*}
    \centering
    \includegraphics[width=16.9cm]{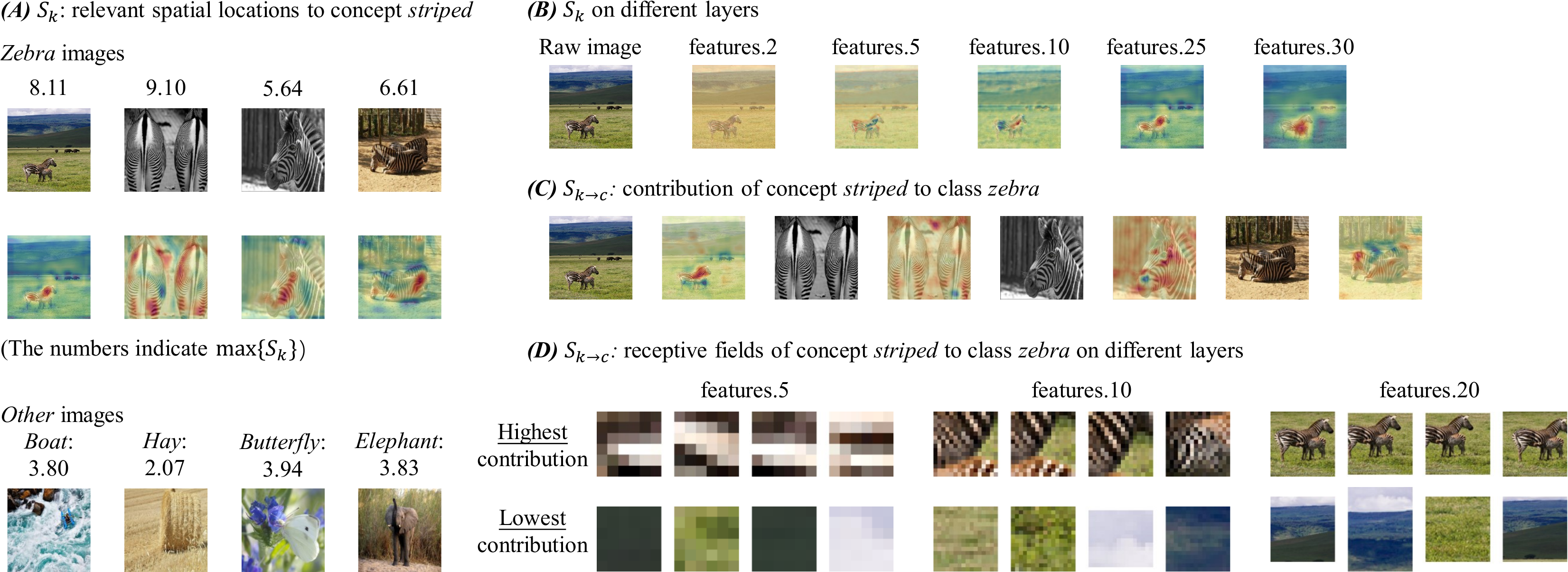}
    \setlength{\belowcaptionskip}{0pt}
    \caption{Experiment results. \textbf{(A\&B)} Results of explanation map $\mbox{\boldmath$S$}_{k}$ which indicates the relevance of each spatial location to concept $k$. \textbf{(C\&D)} Results of explanation map $\mbox{\boldmath$S$}_{k \rightarrow c}$ which quantifies the contribution of concept $k$ to model prediction $c$. See Sec. \ref{sec:exp} for details.}
    \label{fig:results}
\end{figure*}

\section{Method}
\label{sec:method}

Given deep network $f$, chosen layer $l$, input image $\mbox{\boldmath$x$}$ with target class $c$, and query concept $k$, our goal is to calculate the contribution of concept $k$ on feature map $f_{l}(\mbox{\boldmath$x$}) \in \mathbb{R}^{C_{l} \times H_{l} \times W_{l}}$ at different spatial activations.

\noindent{\textbf{Query concepts represented by guidance images.}} Given concept set $\mathcal{K}$, for each $t \in \mathcal{K}$, we have guidance images $\{\mbox{\boldmath$z$}_{t}\}$ and assume that all contents of $\mbox{\boldmath$z$}_{t}$ represent concept $t$. Then, we prepare two sets of guidance images: $\{\mbox{\boldmath$z$}_{k}\}$ ($N_{k}$ images and $k \in \mathcal{K}$) which represents concept $k$ and $\{\mbox{\boldmath$z$}_{\mathcal{R}}\}$ ($N_{\mathcal{R}}$ images and $\mathcal{R} \subseteq \mathcal{K} \backslash k$) which represents other random concepts (Fig. \ref{fig:pipeline} (A)).

\noindent{\textbf{Train SACV with guidance images.}} Next, we feed the guidance images to $f$ and divide each feature map to $H_{l} \times W_{l}$ spatial activations (Fig. \ref{fig:pipeline} (B)). After that, as shown in Fig. \ref{fig:pipeline} (C), we train a linear classifier (\ie, a hyperplane) $g_{k}$ to separate the two sets of spatial activations - $\{f_{l}(\mbox{\boldmath$z$}_{k})_{i}\}_{i=1}^{N_{k} \times H_{l} \times W_{l}}$ (concept $k$) and $\{f_{l}(\mbox{\boldmath$z$}_{\mathcal{R}})_{i}\}_{i=1}^{N_{\mathcal{R}} \times H_{l} \times W_{l}}$ (other random concepts). The coefficient vector $v_{k} \in \mathbb{R}^{C_{l}}$ (\ie, the norm vector of the hyperplane) is the \emph{Spatial Activation Concept Vector (SACV)}.

\noindent{\textbf{Interpretation using SACV.}} As shown in Fig. \ref{fig:pipeline} (D), we first extract input $\mbox{\boldmath$x$}$'s feature map $f_{l}(\mbox{\boldmath$x$})$ and gradient map $\triangledown f_{l}^{c}(\mbox{\boldmath$x$})$ (with respect to target class $c$). In Fig. \ref{fig:pipeline} (E), \textbf{1)} to indicate the relevance of each spatial activation to concept $k$. we multiply each spatial activation with SACV, noted as $f_{l}(\mbox{\boldmath$x$})_{i,j}^{T}{v_{k}}$. There, we obtain a explanation map $\mbox{\boldmath$S$}_{k} = [f_{l}(\mbox{\boldmath$x$})_{i,j}^{T}{v_{k}}] \in \mathbb{R}^{H_{l} \times W_{l}}$. \textbf{2)} Following but being different from TCAV's ``conceptual sensitivity'' score, we calculate concept $k$'s contribution to class $c$ for each spatial activation. We obtain another explanation map $\mbox{\boldmath$S$}_{k \rightarrow c} = [\triangledown f_{l}^{c}(\mbox{\boldmath$x$})_{i,j}^{T}{v_{k}}] \in \mathbb{R}^{H_{l} \times W_{l}}$. Each element in $\mbox{\boldmath$S$}_{k \rightarrow c}$ indicates the contribution of concept $k$ to model prediction of class $c$.

\noindent{\textbf{SACV avoids background interference.}} It is noted that the receptive fields of some $f_{l}(\mbox{\boldmath$x$})_{i,j}$s contain the target object while others do not. As SACV processes spatial activation individually, background interference is removed.

\section{Experiment}
\label{sec:exp}

We test SACV on the PyTorch VGG19 model pertained on ImageNet (the layer names are used in this paper). We use query concepts and guidance images from \emph{Describable Textures Dataset} \cite{cimpoi2014describing}. Results are shown in Fig. \ref{fig:results}.

\noindent{\textbf{$\mbox{\boldmath$S$}_{k}$ indicates the relevance of each spatial location to concept $k$.}} Fig. \ref{fig:results} (A) shows that for \emph{zebra} images, the regions where \emph{zebra} exists have high response to \emph{striped} while for other images, they all have low responses to \emph{striped} for all locations (low $\max\{S_{k}\}$). It means that network $f$ is able to distinguish \emph{striped} from other concepts on this layer. Moreover, Fig. \ref{fig:results} (B) indicates that $f$ recognizes \emph{striped} on high layers (\eg, features.25) but does not learn the concept of \emph{striped} on shallow layers (\eg, features.2).

\noindent{\textbf{$\mbox{\boldmath$S$}_{k \rightarrow c}$ quantifies the contribution of concept $k$ to model prediction $c$.}} As shown in Fig. \ref{fig:results} (C), the red regions in the heat maps indicate the spatial activations which are relevant to concept \emph{striped} while having high contributions to class \emph{zebra}. It is consistent with human understanding as the zebra contents are highlighted. Also, in Fig. \ref{fig:results} (D), the receptive fields of spatial activations show that \emph{striped} features contribute most to model prediction \emph{zebra} while background features have the lowest contributions, meaning that the model makes the right decision with a right reason.


{\small
\bibliographystyle{ieee_fullname}
\bibliography{egbib}
}

\end{document}